\documentclass[11pt,english]{article}

\usepackage{fullpage}
\usepackage[showframe=false]{geometry}
\usepackage{relsize}
\usepackage[T1]{fontenc}
\usepackage[utf8]{inputenc}
\usepackage{xspace}

\usepackage{lineno}
\usepackage{amsmath, amsthm, amssymb, amsfonts, bm}
\usepackage{mathtools}

\DeclarePairedDelimiterX{\infdivx}[2]{(}{)}{%
  #1\;\delimsize|\delimsize|\;#2%
}

\usepackage{enumerate}
\usepackage{xurl}
\usepackage{hyperref}
\hypersetup{
    colorlinks=true,
    linkcolor=blue,
    filecolor=magenta,
    urlcolor=cyan,
}
\usepackage{caption}
\usepackage{subcaption}
\usepackage{breakurl}
\usepackage{graphicx}
\usepackage{booktabs}
\usepackage{tabularx}
\usepackage{multirow}
\usepackage{enumitem}
\usepackage[affil-it]{authblk}
\usepackage{wrapfig}
\usepackage[dvipsnames,svgnames]{xcolor}

\usepackage{natbib}
\setcitestyle{numbers,sort&compress}

\title{Mapping industrial poultry operations at scale with deep learning and aerial imagery}
\author[1,*]{Caleb Robinson}
\author[2]{Ben Chugg}
\author[2]{Brandon Anderson}
\author[1]{Juan M. Lavista Ferres}
\author[2]{Daniel E. Ho}

\affil[1]{Microsoft AI for Good Research Lab, Redmond, WA, 98052}
\affil[2]{Stanford RegLab, Stanford, CA, 94305}
\affil[*]{Corresponding author: \texttt{caleb.robinson@microsoft.com}}
\date{}

\begin{document}
\maketitle

\begin{abstract}
Concentrated Animal Feeding Operations (CAFOs) pose serious risks to air, water, and public health, but have proven to be challenging to regulate. The U.S. Government Accountability Office notes that a basic challenge is the lack of comprehensive location information on CAFOs.
We use the USDA's National Agricultural Imagery Program (NAIP) 1m/pixel aerial imagery to detect poultry CAFOs across the continental United States. We train convolutional neural network (CNN) models to identify individual poultry barns and apply the best performing model to over 42 TB of imagery to create the first national, open-source dataset of poultry CAFOs. We validate the model predictions against held-out validation set on poultry CAFO facility locations from 10 hand-labeled counties in California and demonstrate that this approach has significant potential to fill gaps in environmental monitoring.
\end{abstract}

\section{Introduction}

Concentrated Animal Feeding Operations (CAFOs) are intensive, large-scale industrial farms that, in 2008, produced more than 50\% of the total livestock in the United States~\cite{gurian2008cafos}. The number of CAFOs has continued to increase~\cite{hribar2010understanding}, yet obtaining reliable information on the precise locations and prevalence of CAFOs is difficult, due in part to litigation, limited regulatory capacity, and permit evasion~\cite{handan2021deep}. 
Indeed, the U.S. Government Accountability Office (GAO) noted, ``EPA does not have comprehensive, accurate information on the number of permitted CAFOs nationwide. As a result, EPA does not have the information it needs to effectively regulate these CAFOs''~\cite{us2008concentrated}. 

This lack of oversight gives rise to pressing public health and environmental concerns. A single CAFO can produce more manure than a large city of over 1M people~\cite{us2008concentrated}. CAFOs produce more than 13 times the amount of human waste annually~\cite{burkholder2007impacts}.   
Such waste is often inadequately handled, leading to the contamination of nearby lands and waterways with pathogens~\cite{gerba2005sources}, pharmaceuticals~\cite{campagnolo2002antimicrobial}, heavy metals~\cite{barker1995livestock}, and hormones~\cite{raman2004estrogen}. This causes nitrogen and phosphorous runoffs which severely affect water quality~\cite{kaplan2004manure} and result in algal blooms~\cite{mallin2000impacts}.
CAFOs are also associated with air pollution. Living or attending school in proximity to CAFOs, even at a distance of some miles, is associated with reduced lung function and asthma~\cite{schultz2019residential,sigurdarson2006school,schinasi2011air}.  
Additional public health impacts of CAFOs include propagation and incubation of disease and aggravating climate change~\cite{hribar2010understanding}. Former Secretary of Energy Stephen Chu noted that agriculture and meat production may be more consequential to climate change than power generation~\cite{mcmahon2019}, but such claims are hard to validate without a systematic enumeration of facilities.

These challenges have given rise to recent work on automating CAFO detection and monitoring~\cite{miralha2021spatiotemporal, robinson2021temporal,chugg2021enhancing,handan2019deep}. Handan-Nader and Ho 2019~\cite{handan2019deep} employed a CNN with hand-labeled imagery collected by environmental interest groups over the course of years to automate the mapping of poultry and swine CAFOs across North Carolina. Their approach uses transfer learning with a pretrained network (Inception V3) to classify image tiles. Our work builds on this effort by moving from an image classification to an semantic segmentation framework and focusing on model generalization in order to scale our analysis to the entire U.S. In contrast with a classification approach, the semantic segmentation approach allows us to identify individual barns and thus calculate barn-level features (such as area). Our hypothesis is that an additional filtering step based on barn-level features will improve the performance and generalization ability of our approach (compared to an image classification approach) because industrial poultry production is relatively homogeneous regardless of geography\footnote{Poultry production in the United States operates under the paradigm of ``vertical integration,'' meaning that there are a handful of large companies -- \emph{integrators} -- across the country responsible for contracting out the majority of production to smaller farm operations. Integrators contractually require that production facilities adhere to specific standards. This results in structurally similar poultry barns across the country. See Vukina 2001~\cite{vukina2001vertical} for an overview.}. 

Maroney et al. 2020~\cite{maroney2020using} detect poultry operations in 35 counties across 7 southeastern states, using the ArcGIS Feature Analyst~\cite{blundell2008feature,opitz2008object} on NAIP 1m/pixel aerial imagery. Feature Analyst proceeds by (a) having the user to manually select a training set, (b) using ensembled supervised learning (with a range of models, including neural networks, decision trees, $k$-nearest neighbors), and (c) retraining after the user removes false positives. Patyk et al. 2020 \cite{patyk2020modelling}, augment this ArcGIS-based detection with probabilistic estimates based on the Census of Agriculture \cite{burdett2015simulating} to detect poultry operations in almost 600 U.S. counties.  

These efforts are promising, and our work builds and improves on these efforts in several key ways. First, the modeling approach by Patyk et al. relies on data from the Census of Agriculture (CoA). The CoA, however, is taken only once every 5 years, masks results in small counties, and is based principally on a survey that is affected by non-response and unknown farm locations~\cite{young20172012}. We demonstrate that our approach identifies significant blind spots in the CoA, identifying counties with zero reported CAFOs that, in fact, have large numbers of CAFOs.
Second, the proprietary ArcGIS Feature Analyst system, relies on manual inputs and provides little detail on the exact model deployed for a task; this black box nature makes replication, model verification, and improvements difficult.
Third, while Patyk et al. estimate locations for nearly 600 counties, the resulting data is not publicly available. We hence provide CAFO locations for all 3,000+ counties in the U.S. 

\begin{figure}
\centering
\includegraphics[width=0.8\linewidth]{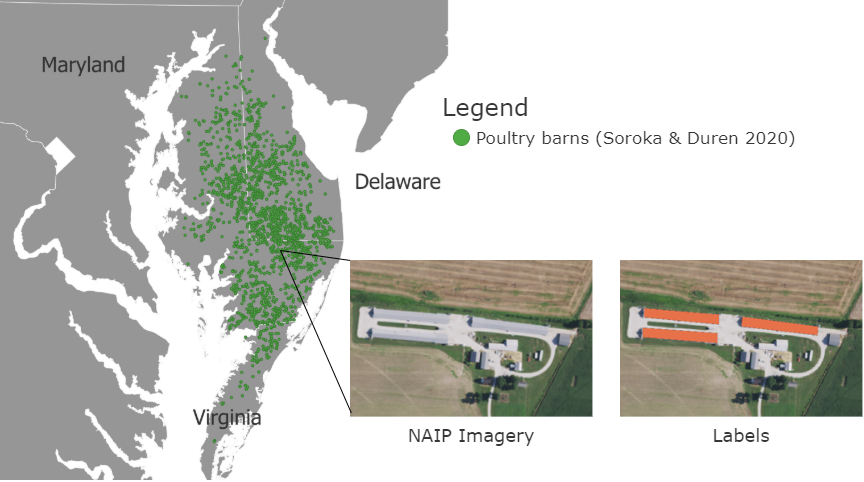}
\caption{Locations of poultry barns from the Delmarva dataset~\cite{sorokaDataset}.}
\label{fig:dataset}
\end{figure}

We take a deep learning approach to provide the first public, comprehensive, national map of poultry CAFOs. Our approach is fully automated, open source, and does not rely on CoA data. 
Specifically, we employ a two-step modeling process. First, we use high-resolution (1m/pixel) four-band (RGB and near-infrared) aerial imagery and poultry barn labels with an open dataset from the Delmarva Peninsula to develop a convolutional neural network (CNN) model. Second, we develop a rule-based filtering methodology to remove false positives predictions made by our model. For instance, predicted poultry barns should follow the distribution of shapes of the labeled barns observed in the Delmarva Peninsula. Similarly, one of the main sources of concept drift is that poultry barns are oriented toward the wind (due to ventilation), and we use rotation augmentation to overcome significant orientation differences across states. The advantage of this approach is that it enables us to leverage the open and high-fidelity Delmarva data when ground truth segmented data can be expensive to acquire in new domains. We validate our model by comparing its predictions to a hand-labeled dataset of poultry CAFOs from ten counties across California, which were not used in the model development. We show that our model is able to obtain recall of 87\% and precision of up to 83\%. 

Our contributions are hence fourfold, providing: 
\begin{itemize}
    \item A method for combining image segmentation with data augmentation and object based filtering to scale CAFO detection using a small set of ground truth labels;
    \item The first national, open-source poultry CAFO dataset with facility locations and size estimates based on building footprints; 
    \item Out-of-sample validation based on large-scale independent labeling by a trained team; 
    \item Results that demonstrate the potential for this approach to detect CAFOs and uncover significant blind spots, for instance, in the Census of Agriculture. 
\end{itemize}
We make all detected facilities, the model, and replication code available at 
\url{https://github.com/microsoft/poultry-cafos/}.

\section{Problem formulation}
\label{sec:problem}
We would like to perform an instance segmentation of remotely sensed imagery to identify individual poultry CAFO barns over the continental U.S. We break this problem in two steps:
\begin{enumerate}[]
    \item a binary semantic segmentation of the remotely sensed imagery into a ``barn'' and ``background'' class and
    \item a supervised grouping of contiguous pixels classified as ``barn'' into barn objects.
\end{enumerate}
Formally, we are given a dataset of $N$ labeled multispectral images and corresponding label masks, $\mathcal{D} = \{(X_i, Y_i)\}_{i=1}^N$.  $X_i \in \mathbb{R}^{h \times w \times c}$ denotes the imagery, where $h$ and $w$ are the pixel height and width, respectively, and $c$ is the number of spectral bands. $Y_i \in \{0, 1\}^{h \times w}$ denotes the label, with 1 denoting a barn and 0 background. 

Step 1 is a standard supervised learning problem in which we aim to learn the parameters, $\theta$, of a semantic segmentation model $f(X_i; \theta) = \hat{Y}_i$ that produces a probabilistic estimate of whether each pixel in an input image is part of the foreground ``barn'' class, $\hat{Y}_i \in \left[0, 1\right]^{h \times w}$. Step  2 is less standard in machine learning, and involves aggregating contiguous groups of predicted foreground pixels into \textit{objects}, $O$, then classifying each of these objects as a ``barn'' or not in terms of their \textit{object-level features}. We engineer or extract $d$ features per object (e.g. the object's area), $O_j \in \mathbb{R}^{d}$, which allows us to incorporate accessory data, besides the multispectral imagery, at the object-level. The classification step can be done through hand-crafted rules, or, for example, by gathering \textit{object-level labels}, $Y_j^\text{object}$, and training a classifier, $g(O_j; \phi) = \hat{Y}^\text{object}_j$.

In the following sections, we describe the datasets we use, the different modeling approaches for the two steps, and experiments to validate our modeling choices and final dataset.

\section{Data} \label{sec:data}

\begin{figure}[t]
    \centering
    \includegraphics[width=0.75\linewidth]{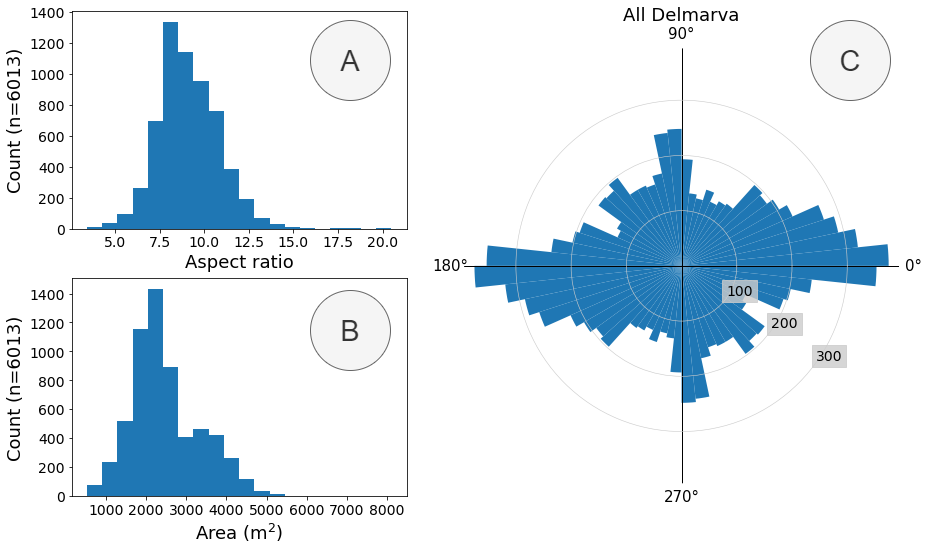}
    \caption{Distributions of properties of the poultry barns in the Delmarva dataset. (\textbf{A}) distribution of the aspect ratio of poultry barns (ratio of the length of the long side of the barn to the length of the short side of the barn) (\textbf{B}) distribution of the area of poultry barns (\textbf{C}) distribution of the orientations of poultry barns.}
    \label{fig:distributions}
\end{figure}

In this work we use aerial imagery from the USDA's NAIP imagery, the Soroka and Duren dataset of poultry barn polygons from the Delmarva peninsula~\cite{sorokaDataset}, a dataset of CAFO facility polygons from an internal validation process, and road network data from Open Street Map.

\paragraph{NAIP Imagery.} The NAIP imagery has a 0.6 to 1m/px spatial resolution, RGB and near-infrared channels, and is collected on a state-by-state basis once every 3 years in the U.S. For consistency, we use the most recent 1m imagery per state, as not all states have transitioned to 0.6m/px imagery. The resulting imagery dataset is substantial in size (>42 TB) and has the advantage of providing comprehensive coverage of the U.S. at high spatial resolution. 

\paragraph{Delmarva Data.} The Delmarva data was produced by the U.S. Geological Survey Chesapeake Bay Studies program to study disease transmission from migratory birds to farms. It consists of 6,013 poultry barn polygons from NAIP 2016/2017 aerial imagery over the Delmarva Peninsula (containing portions of Virginia, Maryland, and Delaware)~\cite{sorokaDataset}. We augment this data using the methodology described by Robinson et al.~\cite{robinson2021temporal} to create estimated construction dates back to 2010/2011 for each poultry barn. Figure~\ref{fig:dataset} depicts the region and instances of the Delmarva data.

\paragraph{Validation Data.} In a separate project examining environmental and public health dimensions of CAFOs in California, we developed an active learning based approach to find CAFO facilities \textit{of any type} or confirm where CAFO facilities exist. We scanned 10 large agricultural counties in California, yielding a validation dataset of 8,869 polygons. Each polygon is labeled as ``empty'' or with the type of CAFO as determined by expert annotators. In contrast to the Delmarva data, the polygon contains the entire facility, including areas that are not the barn. Appendix~\ref{app:validation-data} provides more detail on the data labeling process. Because there are substantial changes in the landscape, surroundings, and adjacent facilities in different regions, we use this validation data as a tough test for our approach to generalize to an agricultural setting that is quite distinct from Delmarva. 

\paragraph{OpenStreetMap Data.} Street data is particularly useful for our object-based filtering for two reasons. First, CAFOs require access to major roads for distribution of livestock or livestock products. Second, roadways can themselves be visually similar to CAFO barns (see Appendix \ref{app:false_positives}), and road networks are hence valuable to filter out false positives. We download all road network location data in the U.S. from OpenStreetMap (OSM) using the \texttt{osmnx} library~\cite{boeing2017osmnx}, and incorporate this data as we describe below. 

\section{Methods}

\begin{figure}[ht]
    \centering
    \includegraphics[width=1.0\linewidth]{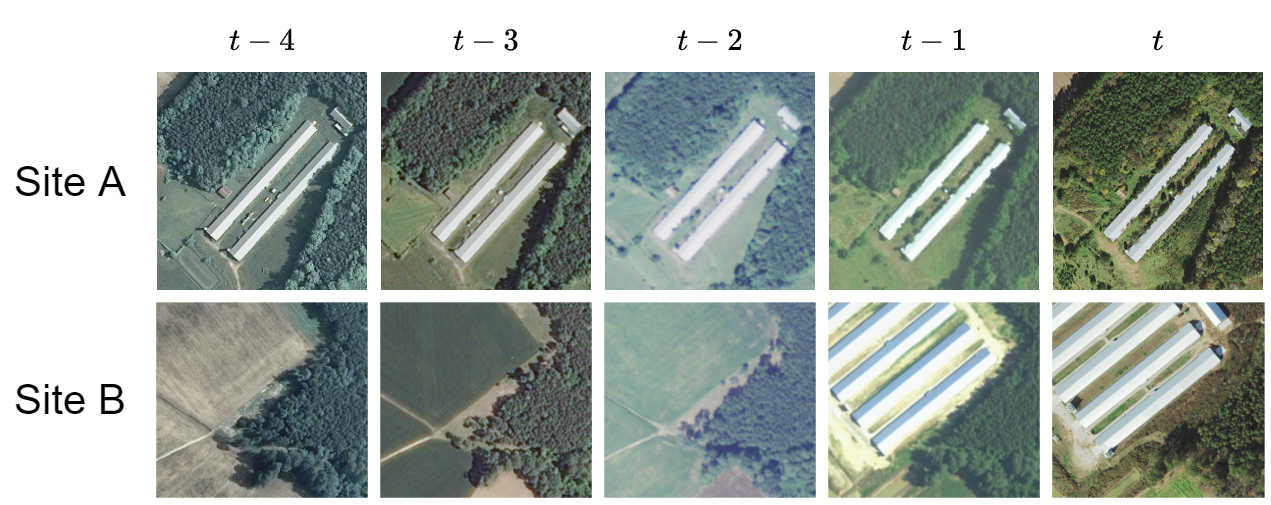}
    \caption{Example of our \textit{temporal augmentation} method. We have labeled data (i.e. pixel masks) and aerial imagery for poultry barns at time $t$, however we also have historical aerial imagery and know many barns will have existed before $t$. With our \textbf{single} training strategy we use imagery and masks from just $t$, with \textbf{all} we use imagery from all time points and assume the same masks apply, and with \textbf{augmented} we use an unsupervised method to determine the previous points in time that the masks are valid for. At ``Site A'' both \textbf{all} and \textbf{augmented} will allow us to train with 4 additional labeled samples in different imaging conditions. However, at ``Site B'' the \textbf{all} strategy will introduce noisy labels, while the \textbf{augmented} strategy will correctly identify that only the imagery at $t-1$ contains a valid mask for training.}
    \label{fig:tcm}
\end{figure}

\subsection{Supervised model training} \label{subsec:supervised_training}
We implement Step 1 from the Problem formulation (Section \ref{sec:problem}) by training U-Net models~\cite{zhou2018unet} to segment NAIP imagery into ``barn'' and ``background'' classes. We use a ResNet-18 encoder as implemented in the \texttt{segmentation-models-pytorch} package~\cite{Yakubovskiy:2019}. We split the Delmarva peninsula spatially into a northern \textit{training area} and a southern \textit{testing area} (roughly 75\% and 25\% of the total area). We further use a $7 \times 6 \text{ km}^2$ area from the training area as a validation set. We do not use a fixed dataset of pre-extracted patches during training, but instead sample patches of size $256 \times 256$ pixels from imagery covering the training area. As only 0.1\% of the training area contains positive ``barn'' masks,\footnote{The training area covers $\sim 14,091\text{km}^2$ while the area of poultry barns in the training area is $13.190\text{km}^2$.} we weight the sampling procedure by randomly discarding a patch with probability $\alpha$ if it does not contain an instance of a positive label. Intuitively, lower values of $\alpha$ result in training on more background pixels, while higher values of $\alpha$ result in training on a more balanced dataset. A value of $\alpha \approx 0.999$ would correspond to a class balanced dataset. 

Given that we only have labeled data from the Delmarva peninsula -- but would like to apply a trained model over the entire U.S. -- our main  concern is that our models will not be exposed to sufficient variation in the data during training and will thus fail to generalize when applied to imagery throughout the country. While poultry barns are relatively standard in appearance regardless of location, their relationship with their surroundings and their orientations can and do vary with location. Further, the NAIP imagery itself varies considerably across states due differences in the time of year, day, and environmental conditions of image capture. For example, Figure \ref{fig:tcm} shows NAIP imagery for two locations at five points in time -- there are considerable non-semantic differences between the images due to the different imaging conditions.

Considering this, we train the models with random rotation augmentation (and random horizontal/vertical flips) to account for shifts in the distribution of poultry barns' orientations. More generally, training with rotation augmentation, or using rotation equivariant networks~\cite{marcos2017rotation,marcos2018land}, is a necessary component of modeling pipelines that use remotely sensed imagery as objects in remotely sensed imagery can be observed in any orientation. We also train with \textit{temporal augmentation} to account for differences in NAIP imagery. This method involves pairing multiple years of NAIP imagery over the training area with a single layer of labels. For example, the Soroka and Duren dataset was created with NAIP imagery from 2016 and 2017 (depending on the state), so without temporal augmentation we would train on a dataset consisting simply of pairs $(X^t, Y^t)$ where $t$ represents the 2016/2017 layer. However, we also have NAIP imagery of the Delmarva peninsula at the same resolution dating back to 2010. With temporal augmentation, we can use the fact that many labels will not change over time, and augment our training set with samples of the form $(X^{t-k}, Y^t)$ for offsets, $k$, that point to valid years of NAIP imagery. While this is a strong assumption, it allows us to expose the model to more variance in the NAIP imagery at the cost of introducing label noise during training. We further use the unsupervised method described in Robinson et al.~\cite{robinson2021temporal} to estimate the construction dates of each poultry barn to avoid augmenting with imagery samples before the construction of the labeled barn. Specifically, we create three versions of our training data, corresponding to different levels of temporal augmentation:
\begin{enumerate}
    \item \textbf{single}, which only uses the 2016/2017 Delmarva labels paired with the corresponding NAIP 2016/2017 imagery
    \item \textbf{all}, which pairs the Delmarva labels with all valid NAIP imagery back to 2010
    \item \textbf{augmented}, which uses estimated construction dates from an unsupervised model to pair the Delmarva labels with appropriate imagery (i.e. in which the labeled poultry barns existed in) back to 2010
\end{enumerate}
See Figure \ref{fig:tcm} for an example of the samples selected by these three approaches. We apply each of these methods to create three training datasets, but keep the test labels fixed to their 2016/2017 imagery. See Table \ref{tab:data_size} for the size of each set.

We train all models with a pixel-wise cross entropy loss, an initial learning rate of 0.01 using the AdamW optimizer~\cite{loshchilov2017decoupled}, and learning rate decay that drops the learning rate by a factor of 10 on training loss plateaus. We compare models trained: with and without filtering (described in the following section); with values of $\alpha \in \{0.05, 0.1, 0.5 \}$; with the three different measures of temporal augmentation (single, all, augmented); and with and without rotation augmentation. In all cases we perform model selection based on validation set performance.

\begin{table}[th]
\centering
\caption{Size of the different training splits (described in Section \ref{subsec:supervised_training}), the testing split, and the full USA imagery (used in Section \ref{subsec:oos_inference}). While the contiguous United States is $\sim8$ trillion $m^2$, the size of the 1m NAIP imagery over the US is $\sim13$ trillion pixels (at a 1m/px resolution) due to overlap between adjacent tiles of imagery.}
\label{tab:data_size}
\begin{tabular}{@{}cccc@{}}
\toprule
\textbf{Split}   & \textbf{Pixels}                       & \textbf{Tiles}              & \textbf{Labeled Barns} \\ \midrule
\textbf{single}  & $\sim$18 billion  & 397   & 5,280  \\
\textbf{augment} & $\sim$124 billion & 1,983 & 24,867 \\
\textbf{all}     & $\sim$124 billion & 1,983 & 26,400 \\
\textbf{test}          & $\sim$5 billion   & 114   & 733    \\ \midrule
\textbf{US-wide imagery} & \multicolumn{1}{l}{$\sim$13 trillion} & \multicolumn{1}{l}{212,354} & -                      \\ \bottomrule
\end{tabular}%
\end{table}

\subsection{Object-based filtering}

The second step of our approach aims to incorporate data, such as a vector road map, that is not possible to directly use in the semantic segmentation model from the first step. Here, we group the per-pixel outputs of the semantic segmentation model to create \textit{objects}, then classify these objects as ``barns'' or not based on object-level features. Specifically, given the probabilistic output of a semantic segmentation model over a large area, $\hat{Y}$, we first threshold the output to create a binary ``barn'' prediction per pixel. The threshold value used can be tuned over a held-out validation set to achieve a desired point on the precision/recall curve. Next, we group contiguous sets of pixels predicted as the positive ``barn'' class into objects (or polygons in geographic space) using a polygonize operation\footnote{A search is performed over predicted pixels to group each contiguous set of pixels into a polygon using the \texttt{rasterio} package\cite{gillies_2019}.} with a 4-pixel neighborhood. We then compute the following set of features for each object/polygon:
\begin{description}
\item[Area] We compute the minimum rotated rectangle~\cite{freeman1975determining} that fits the group of pixels and record the area of this rectangle as the area of the predicted ``barn''. A minimum rotated rectangle for a given polygon is the smallest rectangle with at least one edge coincident to an edge from the polygon. Such a rectangle can be determined in $O(n)$ time where $n$ is the number of vertices in the given polygon~\cite{eberly2015minimum}.
\item[Aspect ratio] We again compute the minimum rotated rectangle and record the ratio of the length of longer side of the rectangle to the length of the shorter side.
\item[Road distance] We compute the distance from each object to the nearest road in the OpenStreetMap database of public and private roads over the U.S.\footnote{For more details about performing this computation at a national scale see Appendix \ref{app:road-distance-features}.}
\end{description}
We use these features to create a rule-based classifier using the \textit{range} of the same features computed over the Soroka and Duren polygons (Figure \ref{fig:distributions}). Any object that has an area outside of the range $\left[525, 8106\right]$ m$^2$, an aspect ratio outside of the range $\left[3.4, 20.49\right]$, or is immediately intersecting with a road in the OpenStreetMap data, is classified as background. In contrast, any object that fits these criteria is classified as a poultry barn. Another way of interpreting this method is as a filtering step that aims to discard groups of false-positive predictions made by the segmentation model. The segmentation model naively operates on image-based data with pixel-level supervision without a direct way to incorporate higher-order labels such as the distribution of aspect ratios of known poultry barns or road network vector data. These features are useful in modeling, and while we used a rule-based classifier with three features in this work, the classifier could be learned using labeled data over a more in-depth object-based feature representation.

\section{Experiments and Results}

\begin{table}[t]
\centering
\caption{Test results of models trained with rotation augmentation.  Results are displayed by (a) whether or not object-level filtering is applied (left and right panels), (b) the level of temporal augmentation (\textbf{single}, \textbf{all}, and \textbf{augmented}), and (c) $\alpha$ weight in the training data. Rows are sorted by the filtered F$_2$ score. Augmentation and $\alpha$ values affect only the training set and evaluation is conducted on the natural distribution of the test set.}
\label{tab:main-results}
\begin{tabular}{@{}cccccccc@{}}
\toprule
\multicolumn{1}{l}{} & \multicolumn{1}{l}{} & \multicolumn{3}{c}{\textbf{Unfiltered}} & \multicolumn{3}{c}{\textbf{Filtered}} \\ \cmidrule(lr){3-5} \cmidrule(lr){6-8}
\textbf{Training set} & $\mathbf{\alpha}$ & Precision & Recall & F$_2$ & Precision & Recall & F$_2$ \\ \midrule
augmented & 0.05 & 27.35\% & 97.25\% & 64.36\% & 87.05\% & 94.68\% & 93.05\% \\
all & 0.1 & 45.64\% & 96.78\% & 79.07\% & 89.94\% & 93.73\% & 92.94\% \\
augmented & 0.5 & 48.90\% & 96.04\% & 80.51\% & 92.01\% & 93.13\% & 92.91\% \\
augmented & 0.1 & 37.74\% & 96.70\% & 73.68\% & 88.51\% & 93.64\% & 92.57\% \\
all & 0.05 & 32.56\% & 96.32\% & 69.21\% & 85.57\% & 93.73\% & 91.98\% \\
all & 0.5 & 36.06\% & 96.61\% & 72.32\% & 87.57\% & 92.45\% & 91.43\% \\
single & 0.1 & 15.95\% & 96.88\% & 48.09\% & 80.60\% & 93.44\% & 90.55\% \\
single & 0.05 & 8.64\% & 98.08\% & 31.93\% & 68.54\% & 93.71\% & 87.30\% \\
single & 0.5 & 43.14\% & 87.71\% & 72.69\% & 86.62\% & 84.30\% & 84.75\% \\ \bottomrule
\end{tabular}
\end{table}

\subsection{Evaluation}

To evaluate a model, we run the model over the entire test area, then group contiguous sets of predicted poultry barn pixels into objects. We consider an object as a true positive prediction if it has greater than a 50\% intersection over union (IoU) with a labeled barn. Similarly, an object is a false positive if it does not have a 50\% IoU with a labeled barn, and a labeled barn is a false negative if no predicted object is counted as a true positive with it. 

We evaluate performance using precision, recall, and the F$_2$ score. We choose the F$_2$ metric to more heavily weight recall in the model evaluation as it is possible to reduce the number of false positives in post-processing, however, it is not possible to reduce the number of false negatives (i.e., find new objects). This point broadly applies to object detection models being run over large amounts of satellite imagery -- there is a much higher cost of missing a potential object-of-interest than there is for making a false positive prediction.

\begin{figure}
    \centering
    \includegraphics[width=1.0\linewidth]{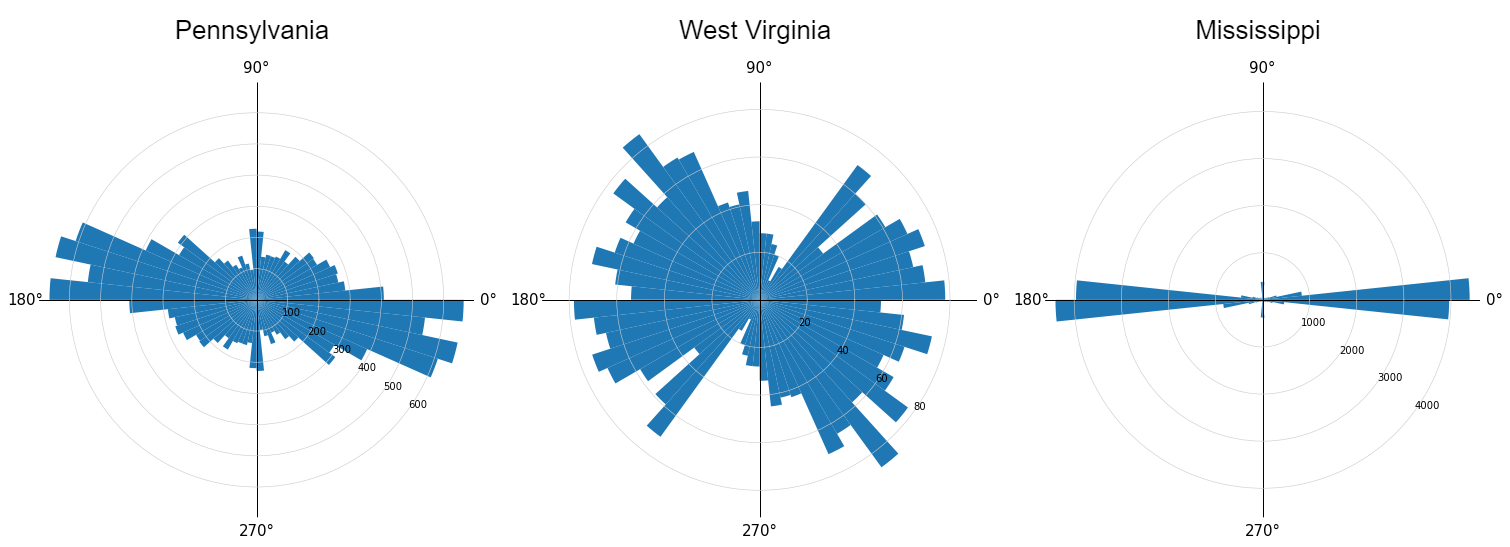}
    \caption{Distributions of orientations of \textit{predicted} poultry barns by state. (\textbf{Left}) Pennsylvania; (\textbf{Middle}) West Virginia; (\textbf{Right}) Mississippi.}
    \label{fig:orientations}
\end{figure}

\subsection{Results}

Table \ref{tab:main-results} summarizes our main results for model variants trained with rotation augmentation.
The Table presents performance measures based on the model's raw predictions (unfiltered columns) and and predictions after applying our proposed object-based filtering method (filtered columns), varying $\alpha$ and temporal augmentation in rows.  Results are sorted by F$_2$ score of filtered models. 

\paragraph{Impact of Filtering.} We find that the filtering step is highly effective based on rules derived from the training set statistics. On average (over the rows of Table \ref{tab:main-results}) recall decreases by 3.29 while precision increases by 52.27. This underscores the substantial utility of expert-based rules in complementing CNN-type approaches. 

\paragraph{Impact of Temporal Augmentation.} We also observe that the U-Net trained with temporal augmentation and an $\alpha$ value of 0.05 performs the best in terms of filtered F$_2$ score. Overall, `augmented' data with lower values of $\alpha$, corresponding to including a higher proportion of background imagery, result in better performance. We also observed in our training process that methods trained with \textit{some} type of temporal augmentation appear to generalize better than those without temporal augmentation -- despite potentially introducing noise into the training process.

\paragraph{Impact of Rotation Augmentation.} We observe a distinct rotation bias in our labeled dataset, where most barns have an East-West orientation (panel C from Figure \ref{fig:distributions}). The reason for this rotation bias lies in ventilation. Amongst growers, one piece of received wisdom is that, ``longest side of your pen, that is, the breadth, should face the prevailing direction of the wind.''\footnote{\url{https://www.justagric.com/poultry-house-construction-guidelines/}}  Of course, wind direction can differ dramatically across regions, so the ability to generalize will be affected by this rotation. Because our training and testing data is from the same area, we do not observe any significant differences between models trained with and without rotation augmentation on our fixed test set. To demonstrate the positive effect of rotation augmentation, we conduct a further experiment with test-time augmentation where we crop a patch of imagery around each poultry barn in the test set, then test each model on all possible 45 degree rotations of these images. The average difference in recall between models trained with and without rotation augmentation is 22.32, while the best model with rotation augmentation shows a 27.47 improvement in recall over its non-rotation augmented counterpart. In other words, if poultry barns are oriented randomly over a landscape, the models with rotation augmentation will identify 22.32\% more poultry barn pixels than the models without rotation augmentation. As expected, we find that other states in the U.S. (outside of the Delmarva peninsula) have different distributions of orientations. For example, Figure \ref{fig:orientations} shows the distribution of orientations of predicted poultry barns in Pennsylvania, West Virginia, and Mississippi.

\subsection{Out-of-Sample Inference and Validation} \label{subsec:oos_inference}

\begin{figure*}[t]
    \centering
    \includegraphics[width=1.0\linewidth]{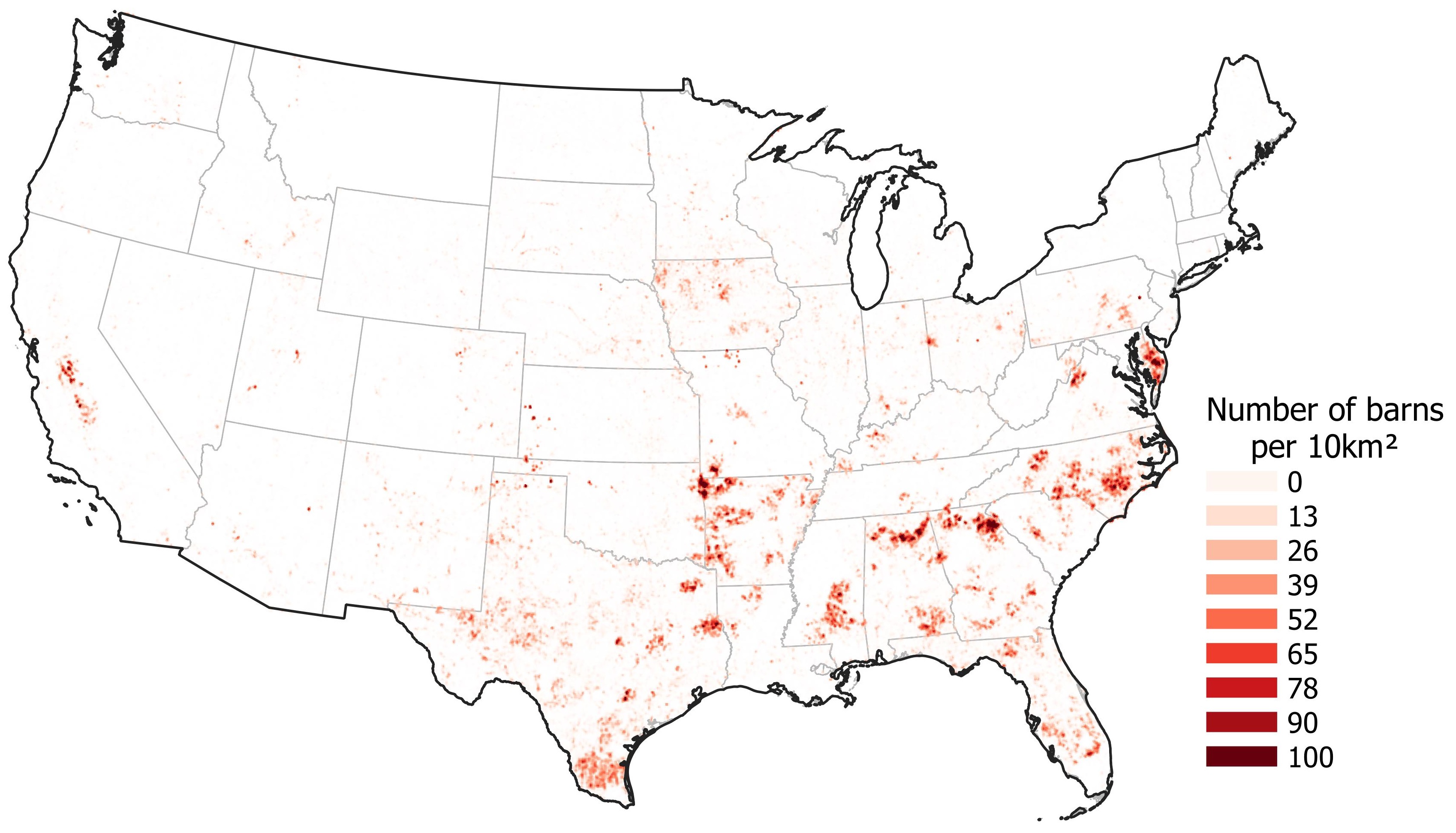}
    \caption{Density rendering of our filtered version of predicted poultry barn locations over the USA. Colors range from 0 barns in a 10km$^2$ radius (transparent) to 100 barns in a 10km$^2$ radius (dark red).}
    \label{fig:usa-map}
\end{figure*}

\paragraph{Inference.} We create a US-wide map of predicted poultry CAFO locations by applying the best performing model over 42 TB of the most recent 1m/pixel NAIP imagery from each state in the continental U.S. This computation was run on Microsoft Azure, specifically with a NC24v3 virtual machine\footnote{The NC24v3 virtual machine type contains 4 NVIDIA V100 GPUs.} that was located in the same cloud region as the NAIP imagery. The NAIP imagery is broken up into $212,354$ \textit{tiles}, each of which is $\sim 8,500 \times 7,000$ pixels, for a total of $\sim13$ trillion pixels. We create a grid of $256 \times 256$ patches with 64 pixels of overlap between neighboring patches from each tile of NAIP imagery, run the model to compute the per-pixel class probability estimates for each patch of imagery, then stitch together the resulting predictions while averaging the class predictions in overlapping areas. The process of averaging predictions along the edges of tiles serves to reduce any border artifacts caused by zero-padding within the model~\cite{huang2018tiling}. We performed this process 4$\times$ in parallel (one process for each GPU on the virtual machine) for a total run time of approximately 2.5 days. This results in 7,108,719 predicted barn polygons before filtering, and 360,857 predicted barn polygons after the filtering step. A density rendering of the predictions are shown in Figure \ref{fig:usa-map}.

\paragraph{Out-of-Sample Validation.} We compare our final filtered set of poultry CAFO barn predictions to the out-of-sample validation dataset described in Section \ref{sec:data} and Appendix \ref{app:validation-data}. This dataset covers 10 counties in California and consists of facility-level annotations (not barn polygons) for \textit{all} types of CAFOs, as well as ``empty'' annotations for polygons that have been confirmed to not have any CAFOs in them.

We count true positives as predictions that are within 100m of a validated poultry facility and false positives as predictions that are not within 100m of any validated poultry facility. For computing recall, we evaluate predictions within the validated area, where false negatives are validated poultry facilities that do not have a prediction within 100m. We apply this methodology over all 10 counties and achieve a recall of 86.90\% and 83.02\% precision.

\section{Implications}
\label{sec:implications}

\begin{figure*}[t]
    \includegraphics[width=1\linewidth]{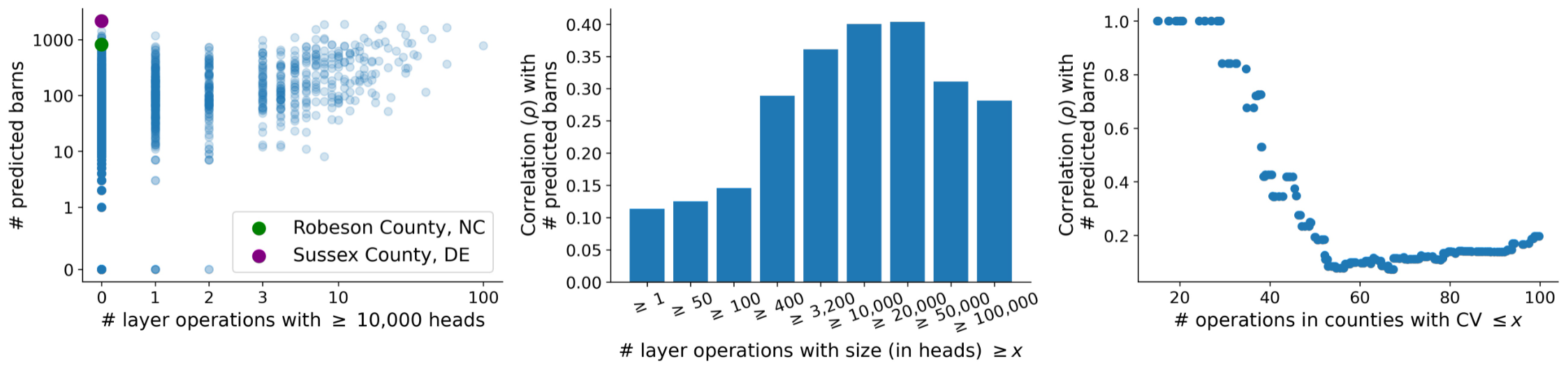}
     \caption{Comparison of our US-wide model predictions with the CoA data at a county level. (\textbf{Left}) Scatter plot showing the number of poultry layer operations with $\geq 10,000$ heads reported by the CoA to the number of predicted barns per county. We highlight Sussex and Robeson counties as extreme examples of where the CoA is missing data (see Figure \ref{fig:sussex_and_robeson}). (\textbf{Middle}) Spearman's correlation coefficient between the number of poultry layer operations over different size categories and our number of predicted barns per county. The correlation is highest for the subset of operations with $\geq 10,000$ heads (i.e. the plot shown in the left panel) indicating that our model is picking up on larger operations. (\textbf{Right}) Spearman's correlation coefficient between the number of poultry operations from subsets of counties with increasing coefficients of variation (CV) as reported by the CoA and our number of predicted barns per county. Our predicted number of barns are highly correlated with the CoA data over the subsets of counties that the CoA data has the most certainty in.}
    \label{fig:coa_comparison}
\end{figure*}

\begin{figure*}[ht]
    \centering
    \includegraphics[width=0.48\linewidth]{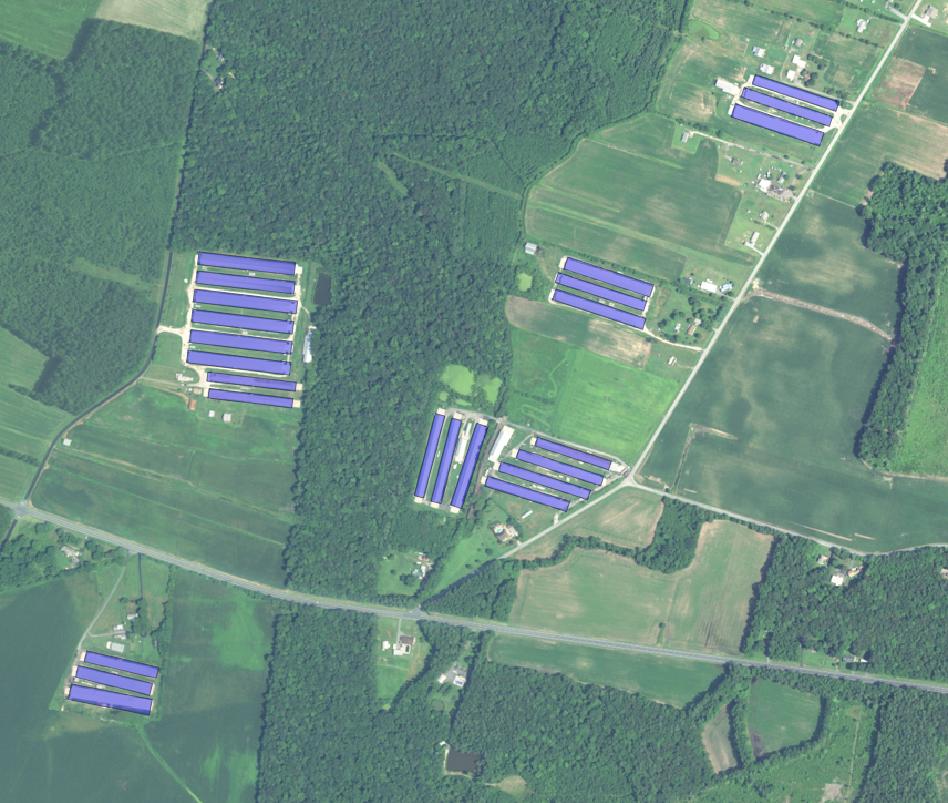}
    \includegraphics[width=0.48\linewidth]{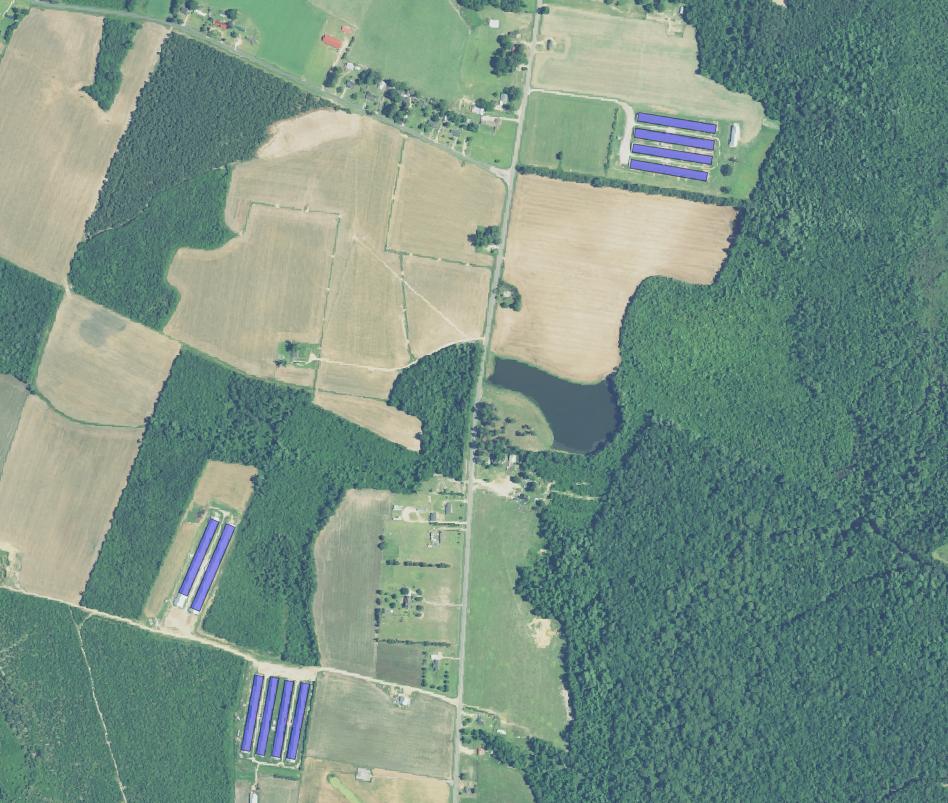}
    \caption{Example model predictions (highlighted in red) from two counties that the CoA reports zero poultry operations in. (\textbf{Left}) 24 predicted poultry barns with NAIP 2017 imagery over a 2.6 km$^2$ area from Sussex County, Delaware; (\textbf{Right}) 10 predicted poultry barns with NAIP 2016 imagery over a 3.1 km$^2$ area from Robeson County, North Carolina.}
    \label{fig:sussex_and_robeson}
\end{figure*}

Our study has substantial implications for environmental research and governance. 

First, we have generated and released the first open-source, national map of poultry CAFOs, which can be used for environmental research, monitoring, and enforcement. Figure \ref{fig:usa-map} shows a heatmap of the filtered version of the predicted poultry barn locations over the U.S. This dataset includes a polygon for each predicted poultry barn with attributes including: the properties used in the filtering step, the average modeled probability over the pixels in the polygon, and the timestamp of the imagery that the predictions were generated from. We have released both sets of filtered and unfiltered predictions for future work to build on. 

Such a map has a myriad of potential use cases. Epidemiological researchers can examine the impact of livestock agriculture of disease spread~\cite{sorokaDataset}. Environmental interest groups can use such information to disseminate and monitor specific facilities~\cite{copeland2010animal}. In 2014, Earthjustice, for instance, filed a petition with EPA about disparate impact of CAFO permitting in North Carolina~\cite{howell2015notification}. Government agencies can use such information to prioritize inspections and understand permitting failures~\cite{purdy2010using,handan2021deep}. For example, the EPA estimated that nearly 60\% of CAFOs do not hold permits~\cite{epa2011}. The national map also enables risk scoring facilities based on the proximity to waterways,  vulnerable communities, and other potential environmental impacts. 

Our approach can also significantly reduce the time currently spent on manually scanning for such facilities by providing a set of likely poultry barns with high-recall. The US contains over 4M km$^2$ of agricultural and pastoral land~\cite{bigelow2017major}. While constructing the out-of-sample validation set we found the task of CAFO identification to be non-trivial even to a human validator, requiring 90 seconds on average for specialists at a third-party company to label a square km image. At this rate we can estimate that it would require approximately 100,000 hours of human effort to cover the US. For states and environmental groups, humans scanning satellite imagery took three or more years to complete a single state~\cite{handan2019deep}. 

Second, our results illustrate major limitations in the Census of Agriculture (CoA), due to  coverage gaps, high nonresponse rates, and, even, internally contradictory data~\cite{mccarthy2018combining,young20172012}. The left panel of Figure~\ref{fig:coa_comparison} plots the detected number of barns by our model per county against the number of operations with over 10,000 heads of layer chickens reported the CoA. While the two are indeed correlated (Spearman's rank order correlation coefficient is 0.38), there are many deviations, particularly in counties where the CoA reports no poultry operations at all. For instance, the CoA lists both Robeson County, North Carolina and Sussex County, Delaware, as having no poultry operations with over 400 heads. In both counties, however, our model detected thousands of poultry barns which we have validated through visual inspection of the NAIP imagery (see Figure~\ref{fig:sussex_and_robeson} for examples from both counties). Both counties also have been previously reported as containing large CAFOs~\cite{graddy2021exposing,rundquist2019under,newman2020here}. 

The middle panel of Figure~\ref{fig:coa_comparison} plots the Spearman's rank order correlation coefficient between the number of barns predicted by our model per county against counts from the CoA based on different poultry operation size thresholds. This shows that the two approaches are most highly correlated with mid- to large-sized CAFOs. Finally, the right panel compares the number of barns predicted by our model to the CoAs count of total operations for subsets of counties based on different uncertainty thresholds reported by the CoA. The CoA reports a coefficient of variation (CV) for each point estimate at the county and state level where smaller values represent less uncertainty, with attempts to adjust for the undercount of facilities~\cite{usda2017}. We collect the subset of counties corresponding to increasing CV value cutoffs, and find that our predictions are highly correlated with the CoA operation counts for counties that the CoA has less uncertainty.

These results show that prior methods to map CAFOs that rely on the CoA may inherit the same coverage gaps. Most promisingly, our approach provides an independent enumeration of CAFOs, without relying on self-reported information by producers. This study also illustrates how advances in computer vision can supplement areas where administrative data have fallen short~\cite{jean2016combining,zhu2016field}. 

Third, our results illustrate advantages to a hybrid approach that leverages segmentation of regular objects (barns) and  expert-based heuristics. Earlier approaches that classify image tiles grapple with centering the CAFO facility in the image and may be sensitive to changes in the background, which could change significantly across U.S. regions (e.g., arid vs. humid landscapes). For instance, the image classification model by Handan-Nader and Ho results in recall of 93.72\% and 71.16\% precision in Delmarva. This is a drop in 16 percentage points in precision. But that model's performance drops significantly when images are not centered around the facility, with recall dropping to 54\%. This suggests that image segmentation is a far more efficient by requiring inference on fewer overlapping image tiles. 

In short, focusing on the key object of the distinct poultry barn and combining expert-based heuristics (e.g., aspect ratio of barns, proximity to road for distribution, and wind orientation of barns) can enable efficient learning with much lower costs in labeling and inference.

\section{Limitations}
\label{sec:limits}

While this study has made a major advance toward an open-source, national poultry CAFO map, we spell out several limitations and opportunities for further work. 

First, the precision of our results varies by geography. From a qualitative error analysis, the most common class of false positive are predictions on roadways that are not covered by OpenStreetMap data and predictions on other linear features that are not included in our current filtering step (e.g. railways). See Appendix \ref{app:false_positives} for examples.
A future avenue of improvement could involve obtaining more comprehensive road data to increase the efficiency of our proximity filter. A tradeoff with this approach, however, is that more comprehensive datasets (e.g. datasets from ESRI or Google Maps) are not open-source. Another alternative would be for researchers to tailor the filtering step to improve the precision based on local knowledge. Future work might also treat the filtering step as an independent learning problem, in which we attempt to learn the distribution of false positives and correct for it. 

Second, while our out-of-sample validation yields strong results, it is not possible to calculate the precision where California data was not labeled.  Since those labels resulted from an active learning model to ensure 80\% recall of facilities --- i.e.. human labelers did not visually inspect 100\% of each county --- it is possible that our precision is biased.  If a predicted polygon is outside of the validated area, we cannot tell whether it is a false positive or a true positive prediction, as it could be a poultry facility that was missed in the validation data.  If we assume that \textit{all} predictions outside of the validated areas are false positives, this yields a extreme lower bound on precision of 18.84\%. The large discrepancy between precision based on overlapping imagery and this lower bound suggests that our model either has a high false positive rate and/or there are missing facilities in the validation data. Qualitatively, we find a mix of both effects, with the model making false positive predictions in desert areas around San Diego, but also identifying likely poultry facilities missed in the validation data. These results suggest that combining sources information will be most promising for a full enumeration of CAFOs. 

Third, our approach has focused specifically on poultry CAFOs and has not examined cattle or hog CAFOs. The reason is that poultry CAFOs are visually distinguishable by the barn, whereas distinguishing hog facilities requires identification of the manure storage system and cattle CAFOs can rely extensively on outdoor feedlots.  A next natural step would be to extend our approach to hog facilities that typically have, for instance, different aspect ratios. 

Fourth, while we demonstrate substantial gains from object-based filtering, there may be other object-level features that can help improve performance. The main challenge here lies in whether such data is comprehensively available across the continental U.S. and validating filters based on such data. For instance, in an earlier iteration, our team attempted to cropland layer data, but, surprisingly, found that such data did not significantly help in classifying CAFOs. Advances in weak supervision may be help to improve the development of such heuristics~\cite{ratner2017snorkel}. 
Last, with the exception of temporal augmentation, this work has not yet examined dynamic effects. It may be harder to directly apply our model to time series of NAIP imagery, as the resolution itself changes, and applicability to more real-time (higher cadence) imagery will require adaptation to lower spatial resolution (see, e.g.,~\cite{chugg2021enhancing}). 

Notwithstanding these limitations, our findings show that this hybrid approach makes a substantial advance in large-scale detection of CAFOs in a transparent, open-source fashion. 

\section{Conclusion}
\label{sec:conclusion}
We have provided the first freely available dataset and corresponding open source model aimed at locating poultry CAFOs across the continental United States. We hope that this work provides an initial step towards improving the regulatory capacity of federal and state environmental agencies ---  filling blind spots in the Census of Agriculture and permit records from state authorities --- as well as the research capacity of academics and the ability of environmental interest groups and the public to monitor these consequential facilities. 

\section*{Acknowledgements}
We thank Microsoft Azure for support in cloud computing, and Schmidt Futures, Stanford Impact Labs, the Chicago Community Trust, and Sarena Snider (Snider Foundation) for research support.

\bibliographystyle{ieeetr}
\bibliography{citations}

\appendix

\section{Validation Data}
\label{app:validation-data}

Using a YOLOv3 model~\cite{redmon2016you} built for a separate project, the inference was run over California to detect CAFOs of any type (swine, dairy, and poultry). 
The predictions were then validated using a team of trained annotators (Stanford undergraduate students), employing an active learning approach based on upper confidence bound (UCB) sampling~\cite{lattimore2020bandit} to reduce the necessary number of images to check. 

Each detected facility is assigned a score in $[1,\infty)$, representing the model's confidence in the prediction. We partition the space $[1,\infty)$ into $K$ discrete buckets $B_1,\dots,B_K$, each bucket $B_i$ associated with an interval $[a_i,b_i)$ where $\cup_i[a_i,b_i)=[1,\infty)$. Bucket $B_i$ contains those images containing a facility with a score in the interval $[a_i,b_i)$. If an image contains multiple detected facilities, it's placed in the bucket associated with its highest scoring facility. We add an additional bucket $B_0$ containing those images with no detected facilities. 

For each bucket $B_i$, let $\mu_i$ denote the fraction of examined images in $B_i$ containing a facility, and let $n_i$ indicate the number of times we've validated an image from this bucket. The UCB score associated with bucket $B_i$ is then 
\begin{equation*}
    S_i = \mu_i + \alpha \left(\frac{\log \sum_j n_j}{n_i}\right)^{1/2},
\end{equation*}
where $\alpha\in\mathbb{R}$ is an ``exploration parameter''. The UCB score reflects the past success of the bucket at containing positive images, but by favoring those buckets which are visited less frequently, the second term ensures that we don't focus exclusively on buckets with the largest percentage of past successes. As $\alpha\to\infty$, we rely less on past success and more on a uniform selection of the buckets. 
To decide which images to sample, we form a probability distribution $\pi$ over the buckets in the natural way: 
\begin{equation*}
    \pi_i = \frac{S_i}{\sum_j S_j}.
\end{equation*}
We sample from $\pi$ with replacement $m$ times, where $m$ is the number of images validated in that round, and then update the scores for the subsequent rounds. For the first round, we take $\pi$ as the uniform distribution. 

This process is repeated in each country until it is estimated that we've found 80\% of the facilities. The estimate is constructed by extrapolating success rates in each bucket over all images. That is, the total number of estimated CAFOs is 
\begin{equation*}
    N_C = \sum_{i=0}^K |B_i|\pi_i.
\end{equation*}
Once we've identified at least $0.8\cdot N_C$ we stop the process and move onto another county. The validation data used in this paper employed this process in 10 counties across California. Note that the process does not result in every image being examined.

\section{Computing road distance features at scale}
\label{app:road-distance-features}

Across the U.S., the OpenStreetMap road network contains $\sim37.8$ million edges\footnote{The number of ``highway'' tags reported by the \url{https://taginfo.geofabrik.de/} web service.}, and our model initially predicted 7.1 million poultry barn polygons. Calculating the distance between each polygon and the closest edge in the road network is thus non-trivial. First, we break the problem up such that it can be solved in parallel by splitting the road network and predictions across the $212,354$ NAIP tiles and only considering nearest matches within a tile (we download the road network aligned to these tile definitions using the \texttt{osmnx} library~\cite{boeing2017osmnx}, and our model predictions are saved at the tile level). Now, we need to compute the shortest distance between $N$ polygons and $M$ lines per tile, however in urban areas both $M$ and $N$ will be prohibitively large (as our model's false positives are often correlated with roads and white buildings) for a naive approach. Our approach is as follows:
\begin{enumerate}
    \item Split each road edge into pieces of at most length $d$. If an edge of length $D$ contained two nodes previously, after this splitting step it would contain $2 + \lfloor D/d \rfloor$ nodes.
    \item Add all nodes from the now split road network into a K-D tree data structure. This data structure offers logarithmic time lookup of the $r$ nearest neighbors of a query point.
    \item For each predicted polygon, query the K-D tree with the polygon's centroid for all neighbors up to a distance of $2d$ away. The set of points returned are guaranteed to include the nodes from the closest line.
    \item Compute the polygon-line distance between each predicted polygon and lines corresponding to the nearest points returned in the previous step. This distance will be the minimum distance.
\end{enumerate}
The road splitting step in this algorithm is necessary to calculate the nearest road line for each given polygon. For example, if we try to find the nearest line for a given polygon based on the endpoints of the line, then we can easily miss long straight roads that pass near to the given polygon. This approach is implemented in the accompanying repository and can process road network data and predictions corresponding to a given tile in seconds.

\section{Examples of false positives} \label{app:false_positives}

\begin{figure}[t]
    \centering
    \includegraphics[width=0.9\linewidth]{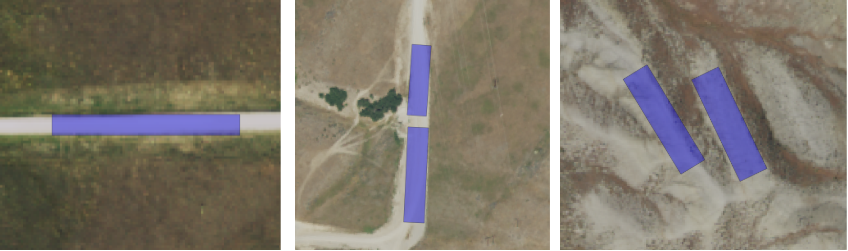}
    \caption{Examples of false positives: (\textbf{left}) sections of white, straight roads in rural South Dakota; (\textbf{middle}) parts of dirt roads in Utah; and (\textbf{right}) barren ground in Wyoming.}
    \label{fig:false_positives}
\end{figure}

Figure \ref{fig:false_positives} shows different types of false positive predictions from three areas in the US. Road false positives are the most common and occur where OSM data is incomplete (e.g. in rural areas, or on informal dirt roads), which breaks the filtering step, or where the OSM roads and the NAIP imagery we use are not aligned. In non-road cases, we observe that the arrangement of the false positives usually mimics a plausible arrangement of barns in a poultry facility (e.g. the pair of false positives shown in the right panel of Figure \ref{fig:false_positives}).

\end{document}